\documentclass[11pt,a4]{article}
\usepackage{eacl2017}
\usepackage{times}
\usepackage{latexsym}
\usepackage{url}
\usepackage{arydshln}
\usepackage{graphicx}
\usepackage{amsmath}
\usepackage{amsfonts, amssymb}
\usepackage{xcolor}
\usepackage[small]{caption}
\usepackage{subcaption}
\usepackage{microtype}

\def\dnrm#1{_{\hbox{\scriptsize #1}}}
\def\dnrmsup#1#2{^#2_{\hbox{\scriptsize #1}}}

\newcommand{\word}[1]{{\em #1}}

\newcommand{\mtag}[1]{{\small{\textsf{#1}}}}
\newcommand{\srctrg}[1]{{\small{\texttt{#1}}}}
\newcommand{\translation}[1]{`#1'}

\usepackage{color}
\usepackage{xcolor}
\definecolor{darkgrey}{rgb}{0.2,0.2,0.2}
\definecolor{grey}{rgb}{0.9,0.9,0.9}
\definecolor{darkblue}{rgb}{0.0,0.0,0.5}
\definecolor{darkred}{rgb}{0.5,0.0,0.0}
\definecolor{darkorange}{rgb}{0.5,0.45,0.4}
\definecolor{darkgreen}{rgb}{0.0,0.6,0.0}
\definecolor{darkyellow}{rgb}{1.0,0.65,0.0}
\definecolor{darkergreen}{rgb}{0.0,0.4,0.0}
\definecolor{lightblue}{rgb}{0.8,0.8,1.0}
\definecolor{lightgreen}{rgb}{0.8,1.0,0.8}
\definecolor{lightred}{rgb}{1.0,0.8,0.8}
\definecolor{lightyellow}{rgb}{1.0,1.0,0.8}
\definecolor{lightorange}{rgb}{1.0,0.9,0.8}
\definecolor{lightgrey}{rgb}{0.96,0.97,0.98}
\definecolor{brilliantlavender}{rgb}{0.96, 0.73, 1.0}

\usepackage{booktabs}
\usepackage{colortbl}
\usepackage{xcolor}
\newcommand{\topheading}[1]{\multicolumn{2}{c}{{\textbf{#1}}}}%

\def\mygray#1{#1}

\long\def\eat#1{}

\title{Neural Multi-Source Morphological Reinflection}

 \author{Katharina Kann \\ CIS \\  LMU Munich, Germany\\ {\tt kann@cis.lmu.de}
          \And  Ryan Cotterell \\ Department of Computer Science \\ Johns Hopkins University, USA \\  {\tt ryan.cotterell@jhu.edu }
          \And Hinrich Sch\"utze \\ CIS \\  LMU Munich, Germany}

\newcounter{notecounter}
\newcommand{\enotesoff}{\long\gdef\enote##1##2{}}

\enotesoff

\def\figref#1{Figure~\ref{fig:#1}}
\def\figlabel#1{\label{fig:#1}\label{p:#1}}
\def\Tabref#1{Table~\ref{tab:#1}}
\def\tabref#1{Table~\ref{tab:#1}}

\def\tablabel#1{\label{tab:#1}\label{p:#1}}
\def\eqref#1{Eq.~\ref{eqn:#1}}

\def\Secref#1{Section~\ref{sec:#1}}
\def\seclabel#1{\label{sec:#1}\label{p:#1}}
\def\secref#1{Section~\ref{sec:#1}}

\definecolor{mylavender}{HTML}{BD71E1}
\definecolor{darkpurple}{HTML}{531B93}

\eaclfinalcopy

\begin{document}

\maketitle
\begin{abstract}

We explore the task of multi-source morphological reinflection, which
generalizes the standard, single-source version. The input consists of (i) a
target tag and (ii) multiple pairs of source form and source tag for a
lemma. The motivation is that it is beneficial to have access to more
than one source form since different source forms can provide
complementary information, e.g., different stems.  We further present
a novel extension to the encoder-decoder recurrent neural architecture,
consisting of multiple encoders, to better solve the task. We show
that our new architecture outperforms single-source reinflection
models and publish our dataset for multi-source morphological
reinflection to facilitate future research.
\end{abstract}

\section{Introduction}
\seclabel{intro}
Morphologically rich languages still constitute a challenge for
natural language processing (NLP). The increased data
sparsity caused by highly inflected word forms in certain languages
causes otherwise state-of-the-art systems to perform worse in standard
tasks, e.g., parsing \cite{ballesteros2015improved} and machine
translation \cite{bojarfindings2}.  To create systems whose performance
is not deterred by complex morphology, the development of NLP tools
for the generation and analysis of morphological forms is crucial.
Indeed, these considerations have motivated a great deal of recent
work on the topic
\cite{ahlberg2015paradigm,dreyer2011non,nicolai2015inflection}.

\begin{table}
\hspace{-.3cm}\begin{tabular}{l@{\quad}l@{\quad}l@{\quad}l@{\quad}l@{\quad}l@{\quad}l}
   & \topheading{Present Ind} & \topheading{Past Ind} & \topheading{Past Sbj}\\
  \specialrule{\heavyrulewidth}{\aboverulesep}{0pt}\arrayrulecolor{black!5}%
  \specialrule{\lightrulewidth}{0pt}{0pt}\arrayrulecolor{black}
   & \mygray{Sg} & \mygray{Pl} & \mygray{Sg} & \mygray{Pl} & \mygray{Sg} & \mygray{Pl}  \\
  \mygray{1} & \textcolor{darkgreen}{tr{\bf e}ffe}         & \textcolor{darkgreen}{tr{\bf e}ffen} & \textcolor{darkpurple}{tr{\bf a}f} & \textcolor{darkpurple}{tr{\bf a}fen} & \textcolor{darkblue}{tr{\bf {\"a}}fe} & \textcolor{darkblue}{tr{\bf {\"a}}fen} \\
  \mygray{2} & \textcolor{darkred}{tr{\bf i}ffst}         & \textcolor{darkred}{tr{\bf e}fft}  & \textcolor{darkpurple}{tr{\bf a}fst} & \textcolor{darkpurple}{tr{\bf a}ft} & \textcolor{darkblue}{tr{\bf {\"a}}fest} & \textcolor{darkblue}{tr{\bf {\"a}}fet} \\
  \mygray{3} & \textcolor{darkred}{tr{\bf i}fft}  & \textcolor{darkgreen}{tr{\bf e}ffen} & \textcolor{darkpurple}{tr{\bf a}f} & \textcolor{darkpurple}{tr{\bf a}fen} & \textcolor{darkblue}{tr{\bf {\"a}}fe} & \textcolor{darkblue}{tr{\bf {\"a}}fen} \\[\jot]
  \bottomrule
\end{tabular}
\caption{The paradigm of the strong German verb {\sc treffen}, which exhibits an irregular ablaut pattern.
  Different parts of the paradigm make use of one of four bolded theme vowels: \textcolor{darkgreen}{{ e}},
  \textcolor{darkred}{{\bf i}}, \textcolor{darkpurple}{{\bf a}} or \textcolor{darkblue}{{\bf {\"a}}}. In a sense,
  the verbal paradigm is partitioned into subparadigms. To see why multi-source models could help in this case, starting
  only from the infinitive \textcolor{darkgreen}{tr{\bf e}ffen} makes it difficult to predict subjunctive form \textcolor{darkblue}{tr{\bf {\"a}}fest},
  but the additional information of the fellow subjunctive form \textcolor{darkblue}{tr{\bf {\"a}}fe} makes the task easier.
}
\label{tab:paradigm}
\end{table}

In the area of generation, the most natural task is morphological
inflection---finding an inflected form for a given target
tag and lemma. An example for English is as follows:
(\srctrg{trg:}\mtag{3rdSgPres}, \word{bring}) $\mapsto$ \word{brings}.  In this
case, the 3rd person singular present tense of \word{bring} is
generated. One
generalization of inflection is morphological {\em re}inflection (MRI)
\cite{cotterell-sigmorphon2016}, where we must produce an inflected
form from a triple of target tag, source form and source
tag. The inflection task  is the special case where the source
form is the lemma. As an example, we may again consider generating the English
past tense form from the 3rd person singular present:
(\srctrg{trg:}\mtag{3rdSgPres}, \word{brought},
\srctrg{src:}\mtag{Past}) $\mapsto$ \word{brings} (where
\srctrg{trg} = ``target tag'' and 
\srctrg{src} = ``source tag''). As the
starting point varies, MRI is more difficult than morphological
inflection and exhibits more data sparsity. However, it is also more widely
applicable since lexical resources are not always complete and, thus,
the lemma is not always available. A more complex German
example is given in \Tabref{paradigm}.

In this work, we generalize the MRI task to a multi-source setup.
Instead of using a single source form-tag pair, we use {\em multiple}
source form-tag pairs.  Our motivation is that (i) it is often
beneficial to have access to more than one source form since different
source forms can provide complementary information, e.g., different
stems; and (ii) in many application scenarios, we will have
encountered more than one form of a paradigm at the point when we want to
generate a new form.

\begin{figure*}
  \captionsetup[subfigure]{aboveskip=15pt,belowskip=-5pt}
  \centering
  \begin{subfigure}{0.245\textwidth}
    \includegraphics[width=\textwidth]{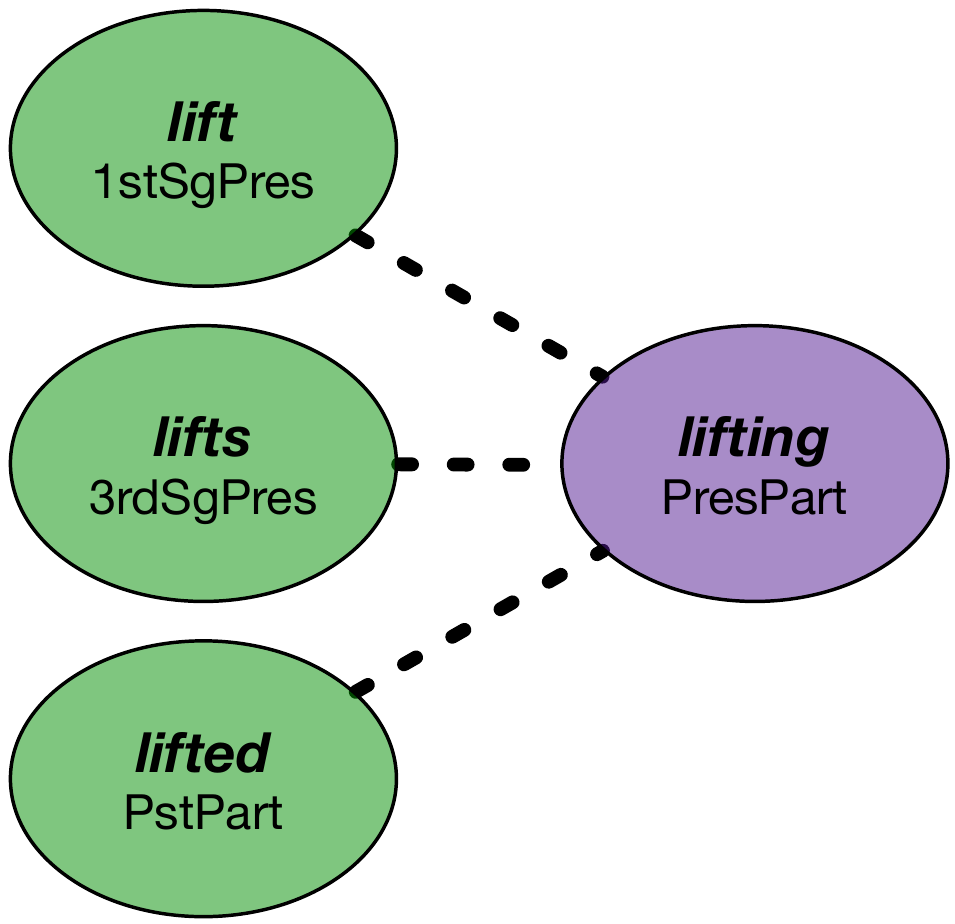}
    \caption{{\sc AnyForm}}
    \label{fig:any}
  \end{subfigure} 
  \begin{subfigure}{0.245 \textwidth}
    \includegraphics[width=\textwidth]{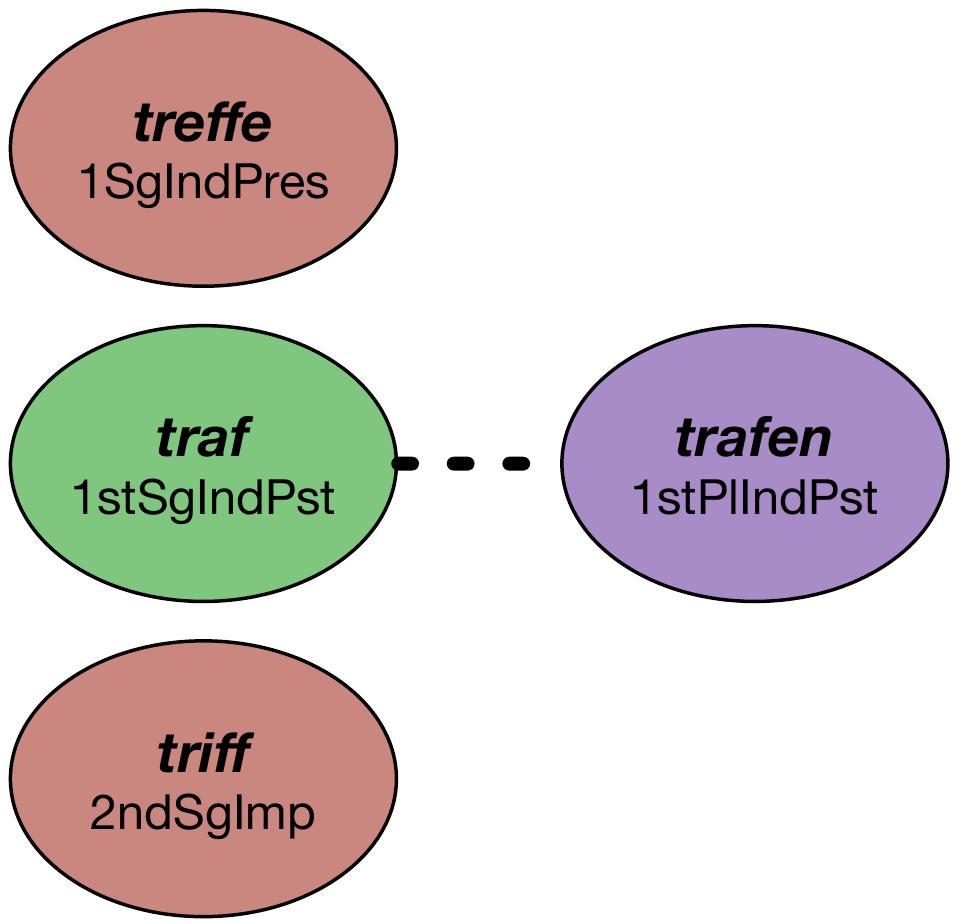}
    \caption{{\sc SingleForm}}
    \label{fig:single}
  \end{subfigure}
  \begin{subfigure}{0.245 \textwidth}
    \includegraphics[width=\textwidth]{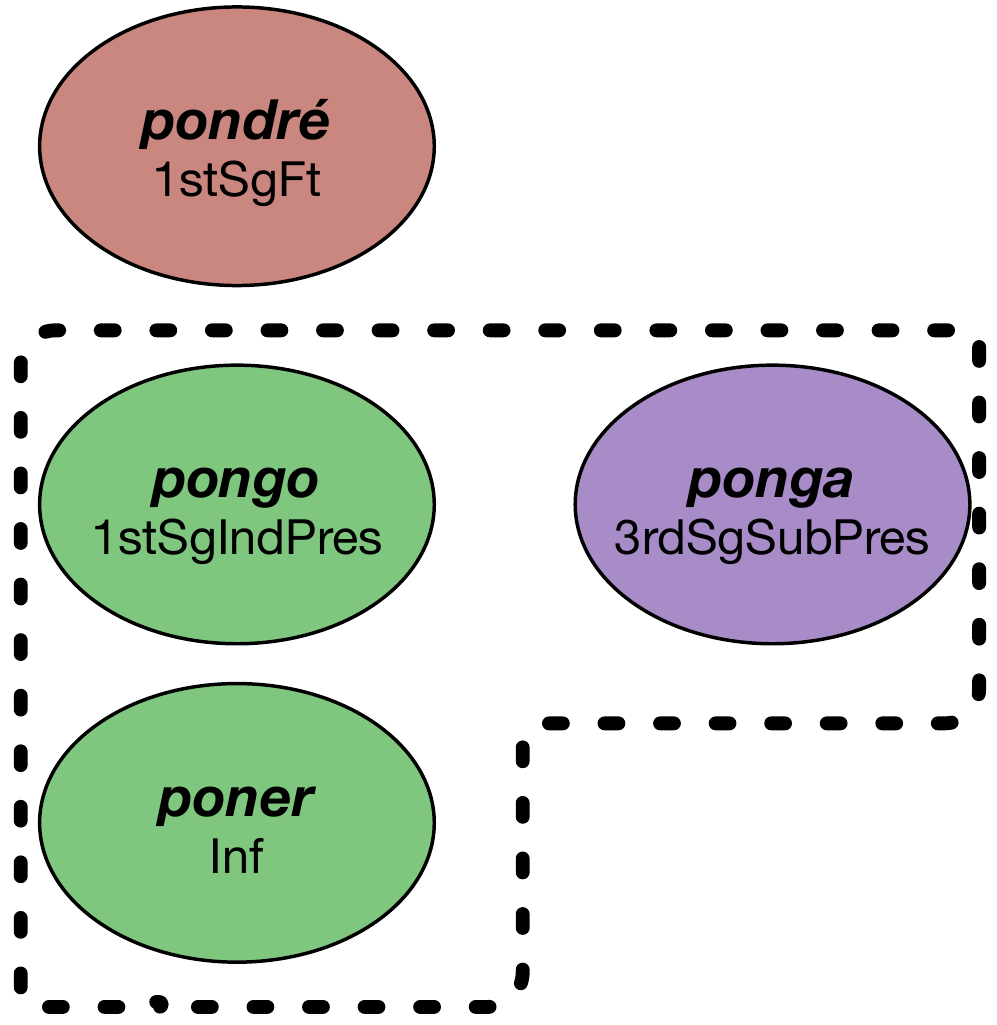}
    \caption{{\sc MultiForm}}
    \label{fig:multi}
  \end{subfigure} 
\begin{subfigure}{0.245 \textwidth}
  \includegraphics[width=\textwidth]{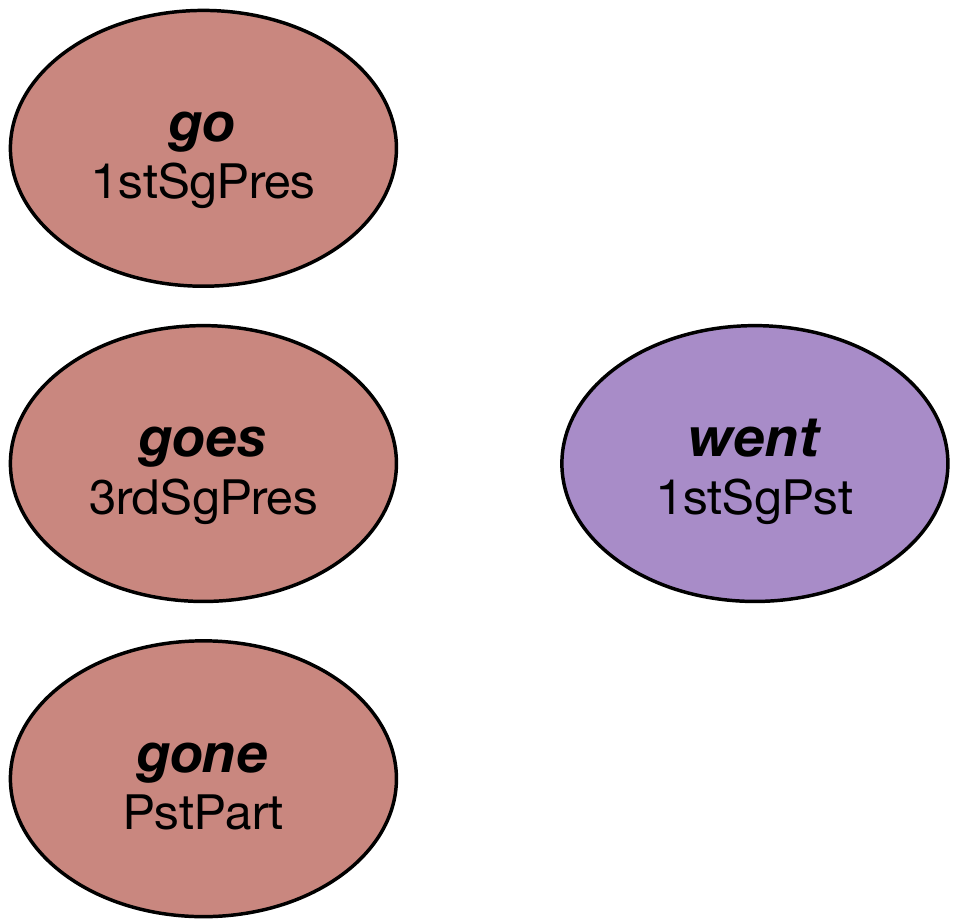}
  \caption{{\sc NoForm}}
  \label{fig:no}
  \end{subfigure}

\caption{Four possible input configurations in multi-source
  morphological reinflection (MRI). In each subfigure, the target form 
  on the right is {\bf \textcolor{darkpurple}{purple}}. The source forms are on the left
  and are {\bf \textcolor{darkgreen}{green}} if they can be used to predict the target form (also connected
  with a dotted line) and {\bf \textcolor{darkred}{red}} if they cannot. There are four possible configurations:
  (i) {\sc AnyForm} is the case where one can predict the target
  form from any of the source forms. (ii) {\sc SingleForm} is the case where
  only one form can be used to regularly predict the target form. (iii) {\sc MultiForm}
  is the case where multiple forms are {\em necessary} to predict the target form. (iv) {\sc NoForm}
  is the case where the target form cannot be regularly derived from any of the source forms.
Multi-source
  MRI is expected to perform better
  than single-source MRI for the configurations \textsc{SingleForm} and
\textsc{MultiForm}, but not for the configurations \textsc{AnyForm} and
 \textsc{NoForm}.
\figlabel{sample}}
\end{figure*}

We will make the intuition that multiple source forms provide
complementary information precise in the next section, but first
return to the English verb \word{bring}. Generating the form \word{brings}
from \word{brought} may be tricky---there is an irregular vowel shift. However,
if we had a second form with the same theme vowel, e.g., \word{bringing}, the
task would be much easier, i.e., (\srctrg{trg:}\mtag{3rdSgPres}, \srctrg{form1:}\word{brought},
\srctrg{src1:}\mtag{Past}, \srctrg{form2:}\word{bringing}, \srctrg{src2:}\mtag{Gerund}).
A multi-source approach clearly is advantageous for this case since 
mapping \word{bringing} to \word{brings} is regular even though the verb itself is irregular.

The contributions of the paper are as follows.  (i) We define the task of
multi-source MRI, a generalization of single-source MRI.  (ii) We show
that a multi-source MRI system, implemented as a novel encoder-decoder,
outperforms the top-performing system in the SIGMORPHON 2016 Shared
Task on Morphological Reinflection on seven out of eight languages,
when given additional source forms. (iii) We release our data to
support the development of new systems for MRI.

\section{The Task: Multi-Source Reinflection}
\seclabel{msr}
Previous work on morphological reinflection has assumed a single
source form, i.e., an input consisting of exactly one inflected source
form (potentially the lemma) and the corresponding
morphological tag. The output is generated from this input. In contrast, 
multi-source morphological reinflection, the task we introduce,
is a generalization in which the model receives multiple
form-tag pairs. In effect, this gives the model a
partially annotated paradigm from which it predicts the
rest.

The multi-source variant is a more natural problem than
single-source morphological reinflection since
we often have access to more than just one form.\footnote{Scenarios where a single form is
available and that form is the lemma are perhaps not infrequent. In high-resource
languages, an electronic dictionary may have near-complete
coverage of the lemmata of the language. However, paradigm
completion is especially crucial for neologisms and low-resource languages.}
For example, corpora
such as the universal dependency corpus
\cite{mcdonald2013universal} that are annotated 
on the token level with inflectional
features often contain several different inflected forms of
a lemma. Such corpora would provide an ideal source
of data for the multi-source MRI task.

Formally, we can think of a morphological paradigm as follows. Let
$\Sigma$ be a discrete alphabet for a given language and $\cal{T}$
be the set of morphological tags in the language. The
inflectional table or morphological paradigm $\pi$ of a lemma $w$ can
be formalized as a set of pairs:
\begin{equation}
\pi(w) =\{ (f_1,t_1), (f_2, t_2), \ldots, (f_N,t_N) \},
\end{equation}
where $f_i \in \Sigma^+$ is an inflected form of $w$, and $t_i \in \cal{T}$ is the
morphological tag of the form $f_i$. The integer $N$ is the number of
slots in the paradigm that have the syntactic category (POS) of $w$.

Using this notation, single-source morphological reinflection (MRI)
can be described as follows. Given a target tag and a pair of source
form and source tag $(t\dnrm{trg},(f\dnrm{src}, t\dnrm{src}))$ as
input, predict the target form $f\dnrm{trg}$.  There has been a
substantial amount of prior work on this task, including systems that
participated in Task 2 of the SIGMORPHON 2016 shared task
\cite{cotterell-sigmorphon2016}. Thus, we may define the task of \emph{multi-source morphological
  reinflection} as follows: Given a target tag and a set of $k$
form-tag source pairs $(t\dnrm{trg},\{(f\dnrmsup{src}{1},
t\dnrmsup{src}{1}), \ldots ,(f\dnrmsup{src}{k}, t\dnrmsup{src}{k})\})$
as input, predict the target form $f\dnrm{trg}$.  Note that
single-source MRI is a special case of multi-source MRI for $k=1$.

\subsection{Motivating Examples}
\seclabel{motivate}
\figref{sample} gives examples for four different
configurations that can occur in multi-source
MRI.\footnote{\figref{sample} is not intended as a complete
  taxonomy of possible MRI configurations, e.g., 
there are hybrids of
  \textsc{AnyForm} and \textsc{NoForm} (some forms
  are informative, others are suppletive) and fuzzy variants
(a single form gives pretty good evidence for how to
  generate the target form, but another single form gives
  better evidence).
All of our examples make  additional assumptions, e.g.,
that we have not seen other similar forms in training either
of the same lemma (e.g., \word{poner}) or of a similar lemma
(e.g., \word{reponer}). Hopefully, the
examples are illustrative of the main conceptual
distinction: several single forms each  are sufficient by
themselves (\textsc{AnyForm}), a single, but carefully selected form is
sufficient (\textsc{SingleForm}),
multiple forms are needed
to generate the target (\textsc{MultiForm}) and the target form cannot be predicted (irregular) from the source forms (\textsc{NoForm}).}
We have colored the source forms green and drawn a dotted
line to the target if they contain
sufficient information for correct generation. If two source
forms together are needed, the dotted line encloses both of
them. Source forms that provide no information in the
configuration are colored red (no arrow); note these forms could
provide (and in most cases will provide) useful information
for other combinations of source and target forms.

The first type of configuration is \textsc{AnyForm}: each of the
available source forms in the subset of the English paradigm
(\word{lift}, \word{lifts}, \word{lifted}) contains enough information
for a correct generation of the target form \word{lifting}. The second
configuration is \textsc{SingleForm}: there is a
single form that contains enough information for correct
generation, but it has to be carefully selected.
Inflected forms of the German verb \word{treffen} \translation{to
meet} have different stem vowels (see \Tabref{paradigm}). In
single-source reinflection, producing a target form with one
stem vowel (\word{{a}} in \word{tr{a}fe} in the figure) from a
source form with another stem vowel 
(e.g., \word{e} in \word{treffe})
is
difficult.\footnote{It is not impossible to learn, but
\word{treffen} is an irregular verb, so we cannot easily leverage
the morphology we have learned about other verbs.}

In contrast, the learning problem for the \textsc{Single\-Form}
configuration is much easier in multi-source MRI. The multi-source
model does not have to learn the possible vowel changes of this
irregular verb; instead, it just needs to pick the correct vowel
change from the alternatives offered in the input. This is a
relatively easy task since the theme vowel is identical. So we only
need to learn one general fact about German morphology (which suffix
to add) and will then be able to produce the correct form with high
accuracy. This type of regularity is typical of complex morphology:
there are groups of forms in a paradigm that are similar and it is
highly predictable which of these groups a particular target form for
a new word will be a member of. As long as one representative of
each group is part of the multi-source input, we can select it to generate
the correct form.

In the \textsc{MultiSource} configuration, we are able to use
information from multiple forms if no single form is
sufficient by itself.
For
example, to generate \word{ponga}, \mtag{3rdSgSubPres} of
\word{poner} \translation{to put} in Spanish, we need to know what the stem is
(\word{ponga}, not \word{pona}) and which
conjugation class (\mbox{\word{-ir}}, \word{-er} or \word{-ar}) it is part of
(\word{ponga}, not \word{pongue}). 
The single-source
input \word{pongo}, \mtag{1stSgIndPres}, does not reveal
the conjugation class: it
is compatible with both \word{ponga} and \word{pongue}.
The single-source
input \word{poner}, \mtag{Inf}, does not reveal
the stem for the subjunctive: it is compatible with
both \word{ponga} and \word{pona}---we need both source forms
to generate the correct form \word{ponga}.

Again, such configurations are frequent cross-linguistically, either
in this ``discrete'' variant or in more fuzzy variants where taking
several forms together increases our chances of producing the correct
target form. Finally, we call configurations \textsc{NoForm} if the
target form is completely irregular and not related to any of the
source forms. The suppletive form \word{went} is our example for this
case.

\subsection{Principle Parts}
The intuition behind the MRI task draws inspiration from the
theoretical linguistic notion of {\bf principle parts}
\cite{finkel2007principal,stump2013morphological}. The notion is that
a paradigm has a subset that allows for maximum predictability. In
terms of language pedagogy, the principle parts would be a minimial
set of forms a student has to learn in order to be able to generate any
form in the paradigm.  For instance for the partial German paradigm in
\Tabref{paradigm}, the forms \word{\textcolor{darkgreen}{treffen}},
\word{\textcolor{darkred}{trifft}}, \word{\textcolor{darkpurple}{trafen}},
and \word{\textcolor{darkblue}{tr{\"a}fen}} could form {\em one} potential set of principle
parts.

From a computational learning point of view, maximizing
predictability is always a boon---we want to make it as easy as possible
for the system to learn the morphological regularities and
subregularities of the language. Giving the system the
principle parts as input is one way to achieve this.

\section{Model Description}
\seclabel{ModelDescription}

Our model is a multi-source extension of MED,
\newcite{kann16singlemodel}'s encoder-decoder network for MRI.  In
MED, a single bidirectional recurrent neural network (RNN) encodes the
input.  In contrast, we use multiple encoders to be able to handle
multiple source form-tag pairs.  In MED, a decoder RNN produces the
output from the hidden representation. We do not change this part of
the architecture, so there is still a single decoder.\footnote{The
  edit tree \cite{chrupala2008towards,thomasjoint} augmentation
  discussed in \newcite{kann16singlemodel} was not employed here.}

\subsection{Input and Output Format}
For $k$ source forms, our model takes $k$ different inputs of parallel
structure.  Each of the $1 \leq i \leq k$ inputs consists of the
target tag $t_{trg}$ and the source form $f_i$ and its corresponding source tag
$t_i$.  The output is the target form.  Each source form is
represented as a sequence of characters; each character is represented
as an embedding.  Each tag---both the target tag and the source tags---is represented as a sequence of subtags; each subtag is represented
as an embedding.

More formally, we define the alphabet $\Sigma\dnrm{lang}$ as the set
of characters in the language and $\Sigma\dnrm{subtag}$ as the set of
subtags that occur as part of the set of morphological tags $\cal{T}$
of the language, e.g., if \mtag{1st\-Sg\-Pres} $\in \cal{T}$, then
\mtag{1st}, \mtag{Sg} and \mtag{Pres} $\in \Sigma\dnrm{subtag}$.  Each
of the $k$ inputs to our system is of the following format:
$S\dnrm{start} \Sigma^+\dnrm{subtag} \Sigma^+\dnrm{lang}
\Sigma^+\dnrm{subtag} S\dnrm{end}$ where the first subtag sequence is
the source tag $t_i$ and the second subtag sequence is the target tag.
The output format is: $S\dnrm{start} \Sigma^+\dnrm{lang} S\dnrm{end}$,
where the symbols $S\dnrm{start}$ and $S\dnrm{end}$ are predefined start and end
symbols.

\begin{figure*}
  \includegraphics[width=\textwidth]{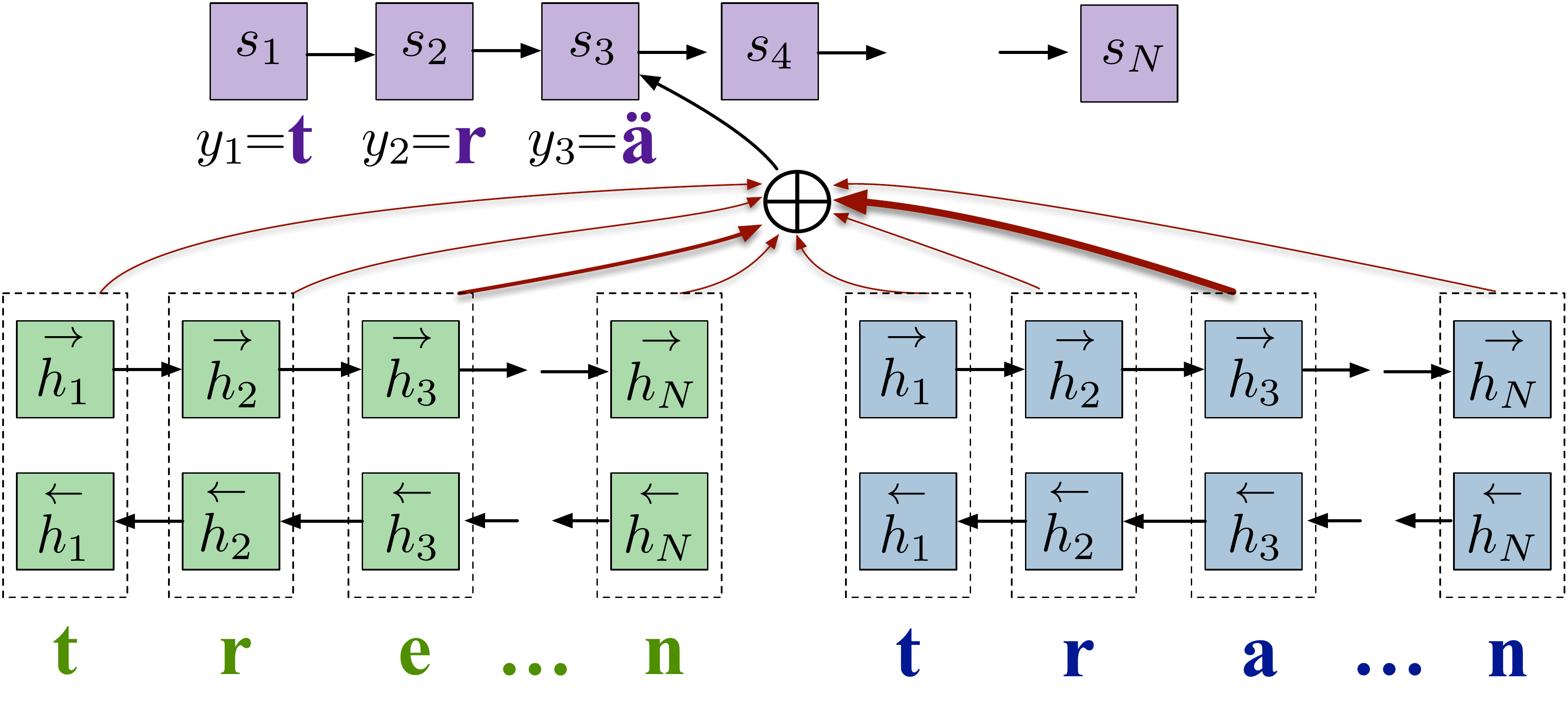}
  \caption{Visual depiction of our multi-source encoder-decoder RNN. We sketch
    a two encoder model, where the left encoder reads in the present form \textcolor{darkgreen}{tr{\bf e}ffen}
    and the right encoder reads in the past tense form \textcolor{darkblue}{tr{\bf a}fen}. They work together
    to predict the subjunctive form \textcolor{darkpurple}{tr{\bf{\"a}}fen}. The shadowed red arcs indicate the strength
    of the attention weights---we see the network is focusing more on \textcolor{darkblue}{{\bf a}} because it helps the decoder
    better predict \textcolor{darkpurple}{{\bf {\"a}}} than \textcolor{darkgreen}{{\bf e}}. We omit the source and target tags as input for conciseness.}
  \label{fig:model}
\end{figure*}

\subsection{Multi-Source Encoder-Decoder}
The encoder-decoder is based on the machine translation model of
\newcite{bahdanau2014neural} and all specifics of our model are
identical to the original presentation unless stated otherwise.\footnote{We
modify the implementation of the model freely available
  at \url{https://github.com/mila-udem}.}
Whereas 
\newcite{bahdanau2014neural}'s model has only one encoder, our model
consists of $k\geq1$ encoders and processes $k$ sources
simultaneously.  The $k$ sources have the form $X_m = (t\dnrm{trg},
f\dnrmsup{src}{m}, t\dnrmsup{src}{m})$, represented as $S\dnrm{start}
\Sigma^+\dnrm{subtag} \Sigma^+\dnrm{lang} \Sigma^+\dnrm{subtag}
S\dnrm{end}$ as described above. Characters and subtags are
embedded.

The input to encoder $m$ is $X_m$.
Each encoder consists of a
bidirectional RNN that computes a hidden state $h_{mi}$ for
each position, the concatenation of forward and backward hidden states.
Decoding proceeds as follows:
 \begin{align}
\label{eq:2}
  p(y \mid X_1,\ldots,X_k) &= \prod_{t=1}^{|Y|} p(y_t \mid \{y_1, ..., y_{t-1}\},
  c_t)  \nonumber \\
       &= \prod_{t=1}^{|Y|}g(y_{t-1}, s_t, c_t),
\end{align}
where $y = (y_1, ..., y_{|Y|})$ 
is the output sequence (a sequence of $|Y|$ characters),
$g$ is a nonlinear function, $s_t$ is the hidden state of
the decoder and $c_t$ is the sum of the encoder states $h_{mi}$,
weighted by attention weights $\alpha_{mi}(s_{t-1})$ that depend on the decoder state:
\begin{equation}
c_t = \sum_{m=1}^{k}\sum_{i=1}^{|X_m|} \alpha_{mi}(s_{t-1}) h_{mi}.
\end{equation}
A visual depiction of this model may be found in \figref{model}.
A more complex hierarchical attention structure would be an
alternative, but this simple model in which all hidden
states contribute on the same level in a single attention layer
(i.e.,
$\sum_{m=1}^{k}\sum_{i=1}^{|X_m|} \alpha_{mi} = 1$) works
well as our experiments show. The $k$ encoders share their weights.

\section{Multi-Source Reinflection Experiment}
\seclabel{Experiments}

We evaluate the performance of our model in an experiment based on
Task 2 of the SIGMORPHON Shared Task on Morphological Reinflection
\cite{cotterell-sigmorphon2016}.  This is a single-source MRI
task as outlined in \Secref{intro}.

\subsection{Experimental Settings}

\paragraph{Datasets.} 
Our datasets are based on the data from the SIGMORPHON 2016
Shared Task on Morphological Reinflection
\cite{cotterell-sigmorphon2016}.  Our experiments cover
eight languages: Arabic, Finnish, Georgian, German,
Hungarian, Russian, Spanish and
Turkish. The languages were chosen to represent
different types of morphology. Finnish, German, Hungarian,
Russian, Turkish and Spanish are all suffixing. In addition
to being suffixing, three  of these languages employ
vocalic (German, Spanish) and consonantal (Russian)
stem changes for many inflections.
The members of
the remaining sub-group are agglutinative.  Georgian makes use of
prefixation as well as suffixation. Arabic morphology
contains both concatenative and templatic elements. We build
multi-source versions of the dataset for Task 2 of the SIGMORPHON
shared task in the following way. We use data
from the {\sc UniMorph}
project,\footnote{\url{http://unimorph.org}} containing
complete paradigms for all languages of the shared task.
The shared task data was sampled from the same set of
paradigms; our new dataset is a superset
of the SIGMORPHON data.

We create our new dataset by uniformly sampling three additional word
forms from the paradigm of each source form in the original data.  In
combination with the source and target forms of the original dataset,
this means that our dataset is a set of 5-tuples consisting each of
four source forms and one target form.\footnote{One thing to note is
  that the original shared task data was sampled depending on word
  frequency in unlabeled corpora. We do not impose a similar
  condition, so the frequency distributions of our data and the shared
  task data are different. Also, we excluded Maltese and Navajo due to a lack of data to
  create the additional multi-source datasets.}
\begin{table}
\centering
  \begin{tabular}{l|rrrr}
    & {1} & {2}  & {3} & \multicolumn{1}{c}{$\geq$ 4}\\ \hline
    ar & 0 & 0 & 0 & 12,800\\  
    fi & 0 & 0 & 0 & 12,800\\  
    ka & 1015 & 84 & 2 & 11,699\\  
    de & 0 & 0 & 0 & 12,800\\  
    hu & 0 & 0 & 0 & 19,200\\  
    ru & 0 & 0 & 5 & 12,794\\  
    es & 1575 & 25 & 877 & 10,323\\  
    tu & 0 & 0 & 0 & 12,800   
  \end{tabular}
  \caption{Number of target forms in the training set for which 1, 2, 3 or
    $\geq$ 4 source forms (in the training set) are available for
    prediction. The tables for the development and test splits show the same
    pattern and are omitted.}
  \tablabel{forms_per_paradigm}
\end{table}
Ideally, we would like to keep the experimental variable
$k$, the number of sources we use in multi-source MRI,
constant for a particular experiment or vary it
systematically across other experimental conditions.
\tabref{forms_per_paradigm} gives an
overview of the
number of different source forms per language in
our dataset. 
Our dataset is available for download at \url{http://cistern.cis.lmu.de}.

\begin{table}
  \centering
  \begin{tabular}{l || c c c c | c | c}
& \multicolumn{6}{c}{source form(s) used}\\
    &  1 &  2 & 3 &  4 & 1--2 & 1--4 \\ \hline\hline
    ar & .871 & .813 & .796 & .830 & .905 & \textbf{.944} \\
    fi & .956 & .929 & .941 & .934 & .965 & \textbf{.978} \\
    ka & .967 & .943 & .942 & .934 & .969 & \textbf{.979} \\
    de & .954 & .922 & .931 & .912 & .959 & \textbf{.980} \\
    hu & \textbf{.992} & .962 & .963 & .963 & .988 & .989 \\
    ru & .876 & .795 & .824 & .817 & .888 & \textbf{.911} \\
    es & .975 & .961 & .963 & .968 & .977 & \textbf{.984} \\
    tu & .967 & .928 & .947 & .944 & .970 & \textbf{.983} 
      \end{tabular}
  \caption{Accuracy on  MRI for single-source (1, 2, 3, 4) and
    multi-source (1--2, 1--4) models.
Best result in bold.}
  \tablabel{multi-source}
\end{table}

\paragraph{Hyperparameters.}
We use embeddings of size 300. Our encoder and decoder GRUs
have 100 hidden units each.  Following \newcite{le2015simple},
we initialize all encoder and decoder weights as well as the
embeddings with an identity matrix. All biases are
initialized with zero. We use stochastic gradient descent, Adadelta
\cite{zeiler2012adadelta} and a minibatch size of 20 for
training. Training is done for a maximum number of 90
epochs. If no improvement occurs for 20 epochs, we stop
training early. The final model we run on test is the model
that performs best on the development data.

\paragraph{Baselines.}
For the single-source case, we apply MED, the top-scoring system in
the SIGMORPHON 2016 Shared Task on Morphological Reinflection
\cite{cotterell-sigmorphon2016,kann16singlemodel}.  At the time of
writing, MED constitutes the state of the art on the dataset. For
Arabic, German and Turkish, we run an additional set of experiments to
test two additional architectural configurations of multi-source
encoder-decoders: (i) In addition to the default configuration in
which all encoders share parameters, we also test the option of each
encoder learning its own set of parameters (shared par's: yes vs.\ no
in \tabref{architecture_comparison}). (ii) Another way of
realizing a multi-source system is to concatenate all sources and give
this to an encoder-decoder with a single encoder as one
input (encoders: $k=1$ vs.\ $k>1$ in
\tabref{architecture_comparison}).

\paragraph{Evaluation Metric.}
We evaluate on 1-best accuracy (exact match) against the gold form.
We deviate from the shared task, which also evaluates under mean
reciprocal rank and edit distance. We omit the later two  since all these metrics were highly
correlated \cite{cotterell-sigmorphon2016}.

\subsection{Results}
\tabref{multi-source} shows the results of the MRI experiment on test data.  We
compare using a single source, the first two sources and all four
sources.  The first source (in column ``1'') is the original source
from the SIGMORPHON shared task.  Recall that we used uniform sampling
to identify additional forms whereas the sampling procedure of the
shared task took into account frequency. We suspect that this is the
reason for the worse performance of the new sources compared to the
original source; e.g., in German there are rarely used subjunctive
forms like \word{bef\"{a}hle} that are unlikely to help
generate related forms that are more frequent.

The main result of the experiment is that multi-source MRI
performs better than single-source MRI for all languages
except for Hungarian and that, clearly, the more sources
the better: using four sources is always better than using
two sources. This result confirms our hypothesis,
illustrated in \figref{sample}, that for most languages,
different source forms provide complementary information
when generating a target form and thus performance of the
multi-source model is better than of the single-source model.
\tabref{multi-source} demonstrates that 
the two configurations we identified as
promising for multi-source MRI, 
\textsc{SingleForm} and
\textsc{MultiForm}, occur frequently enough to
boost the performance for seven of the eight languages, with
the largest gains observed for Arabic (7.3\%) and 
Russian (3.5\%) and the smallest for Spanish
(0.9\%) and Georgian (1.3\%) (comparing using source
form 1 with using source forms 1--4).

\begin{table}
\centering
\begin{tabular}{l||l c | cc}
    { encoders:} &  &\multicolumn{1}{c|}{$k=1$} &\multicolumn{2}{c}{$k=4$}\\
    { par's shared:} &    && \multicolumn{1}{c}{yes} &\multicolumn{1}{c}{no}  \\ \hline \hline
    &ar   &\textbf{.944} & \textbf{.944} & .920 \\
 &de   & \textbf{.980}  &  \textbf{.980} & .975 \\
    &tu  &\textbf{.985}   & .983 & .969 
  \end{tabular}
  \caption{Accuracy of different architectures for the
    dataset with 4 source forms
    being available for prediction. The best result for each row is in bold.}
  \tablabel{architecture_comparison}
\end{table}

Hungarian is the only language for which performance
decreases, by a small amount (0.3\%). We attribute this to overfitting: the
multi-source model has a larger number of parameters,
so it is more prone to overfitting. We would expect the
performance to be the same in a comparison of two models
that have the same size.

\begin{figure*}
  \centering
  \begin{subfigure}[b]{0.32\textwidth}
    \includegraphics[width=\textwidth]{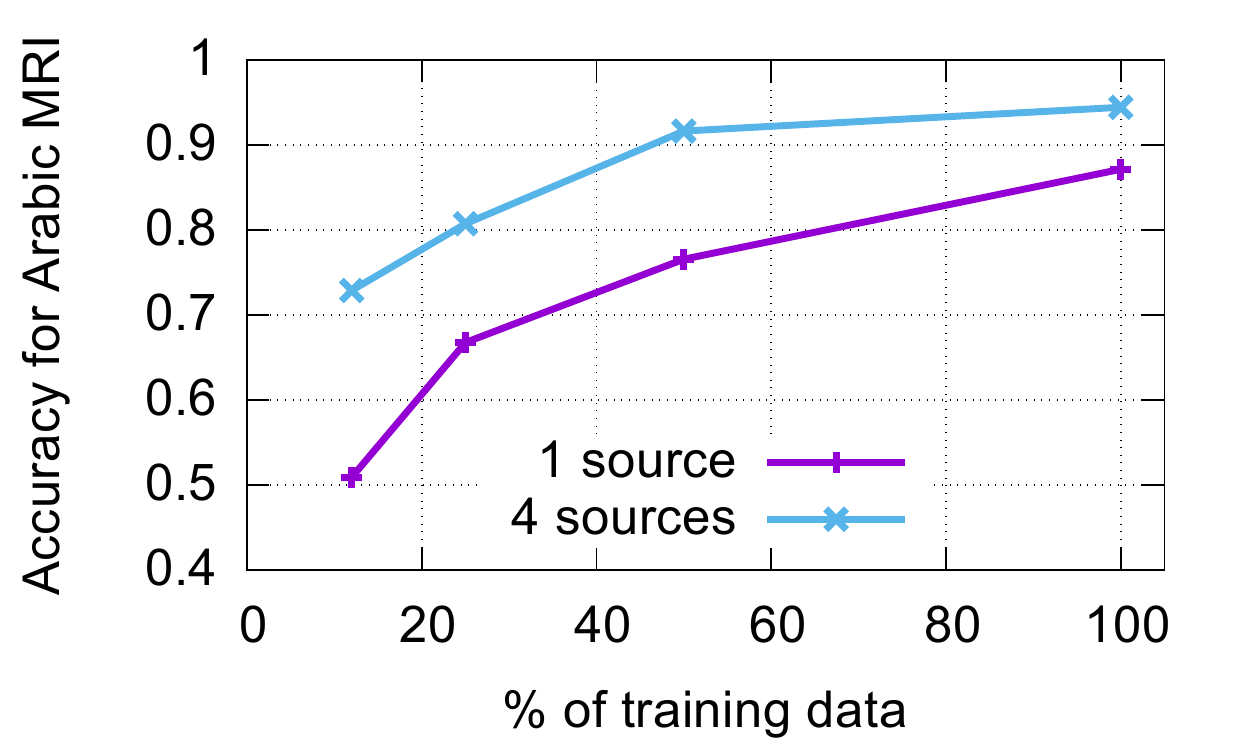}
    \caption{Arabic}
    \label{fig:arabic}
  \end{subfigure} 
  \begin{subfigure}[b]{0.32 \textwidth}
    \includegraphics[width=\textwidth]{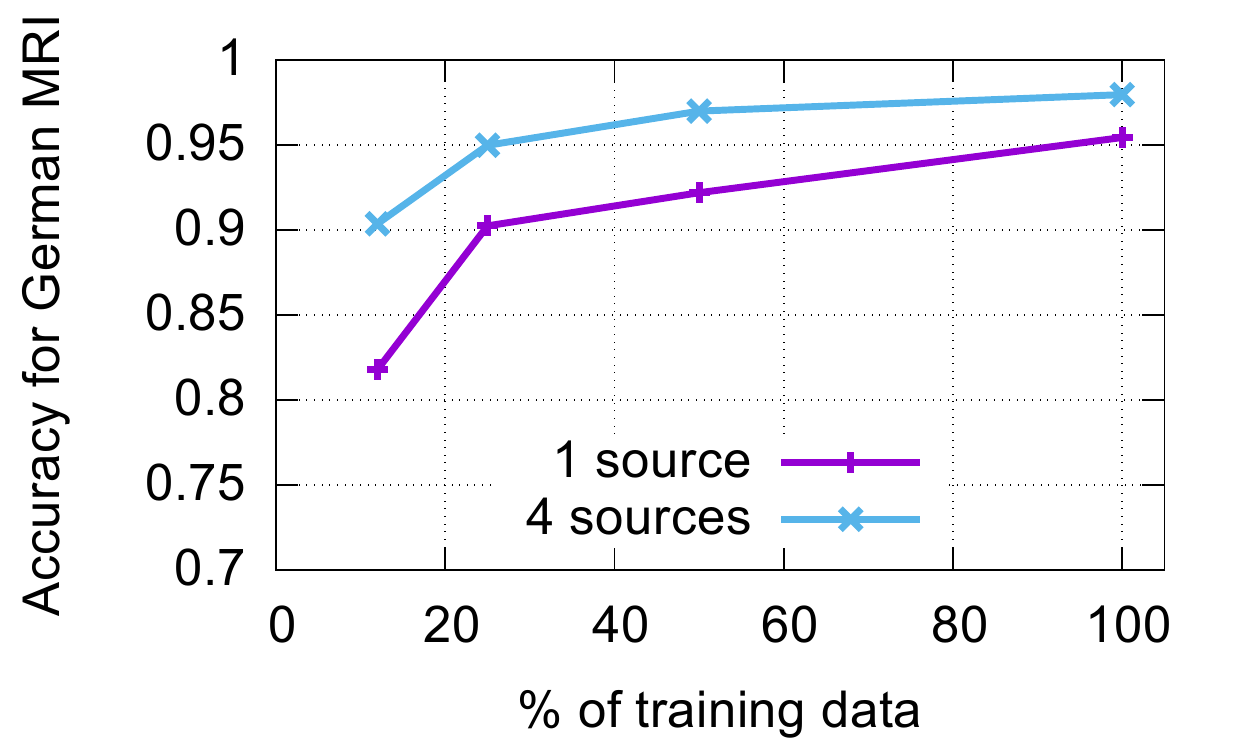}
    \caption{German}
    \label{fig:german}
  \end{subfigure}
  \begin{subfigure}[b]{0.32 \textwidth}
    \includegraphics[width=\textwidth]{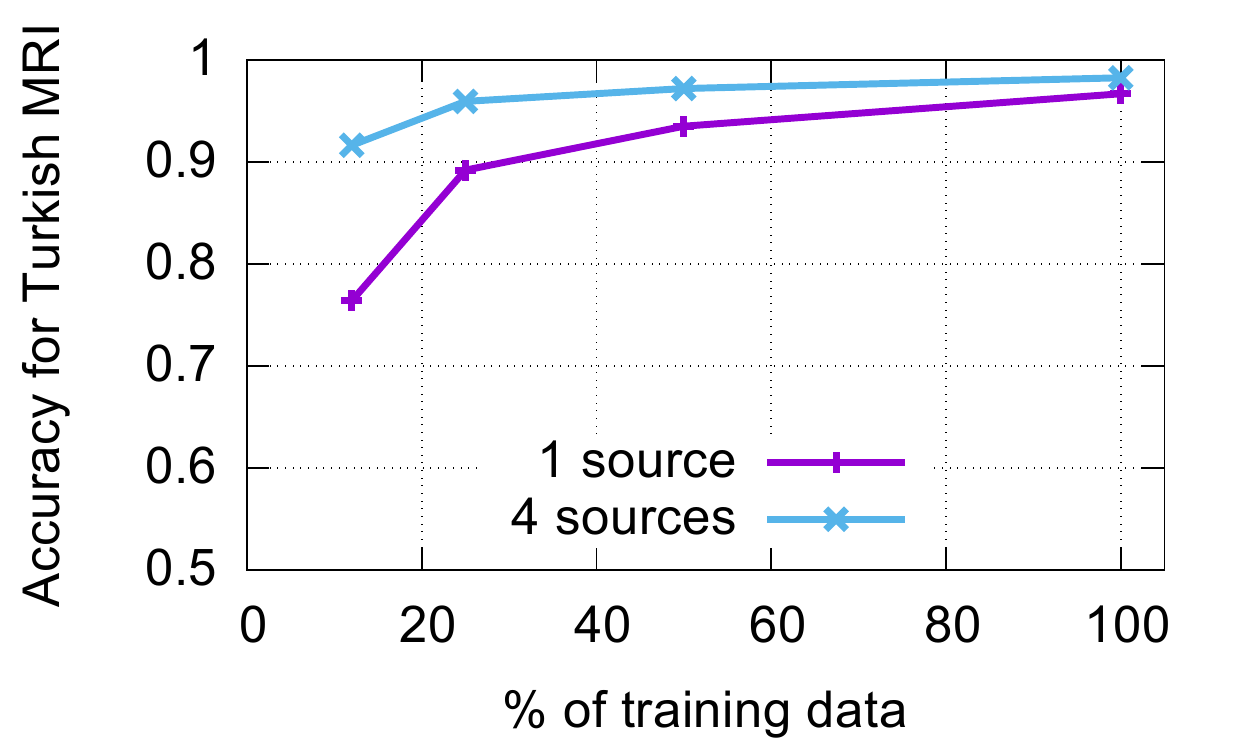}
    \caption{Turkish}
    \label{fig:turkish}
  \end{subfigure}
  
  \caption{Learning curves for single-source and
    multi-source models for Arabic, German and Turkish\label{fig:ana_12}. We observe
    that the multi-source model generalizes faster than the single soure case---this
  is to be expected since the multi-source model often faces an easier transduction problem. }
\end{figure*}

\paragraph{Error Analysis.}
We compare errors of single-source and multi-source models for German
on development data. Most mistakes of the multi-source model are stem-related:
\word{versterbst} for
\word{verstirbst}, \word{erwerben} for
\word{erw\"urben},
\word{Apfelsinenbaume} for
\word{Apfelsinenb\"aume},
\word{lungenkr\"ankes} for
\word{lungenkrankes} and 
\word{\"ubernehmte}
for \word{\"ubern\"ahme}. In most of these cases, the stem of the
lemma was
used, which is correct for some forms, but not for the form
that had to be generated.
In one case, the multi-source model did not use
the correct inflection rule:
\word{braucht} for
\word{gebraucht}---the inflectional rule that the past
participle is formed by \word{ge-} was not applied.

Errors of the
single-source model that were ``corrected'' by the
multi-source model include
\word{empfahlt} for \word{empfiehl},
\word{Throne} for 
\word{Thron} and \word{befielen} for
\word{befallen}. These are all \textsc{SingleForm} cases:
the multi-source model will generate the correct form if it
succeeds in selecting the most predictive source form. The
single-source model is at a disadvantage if this most
predictive source form is not part of its input.

\subsection{Comparison of Different Architectures} 
\tabref{architecture_comparison} 
compares different architectural configurations. All
experiments use 4 sources. We see that sharing parameters is
superior as expected. Using a single encoder
on 4 sources performs as well as 4 encoders
(and very slightly better on Turkish). Apparently, it
has no difficulty learning to understand an unstructured (or
rather lightly structured)
concatenation of form-tag pairs; on the other hand, this
parsing task, i.e.,
learning to parse the sequence of form-tag pairs,
is easy, so this is not a surprising result.

\subsection{Learning Curves}
\figref{ana_12} shows learning curves for Arabic, German and
Turkish.  We iteratively halve
the training set and train models for each subset.  In this
analysis, we train all models for 90 epochs, but use the
numbers from the main experiment for the full training set.
For the single-source model, we use the SIGMORPHON source.
The figure shows that the single-source model needs more individual
paradigms in the training data to achieve the same performance as the
multi-source model. 
The largest difference between
single-source and multi-source 
is $>$ 20\% for Arabic when only 1/8 of the training set is used.
This suggests that multi-source MRI is an attractive
option for low-resource languages since it exploits available data better than single-source.

\begin{figure*}
\includegraphics[width=1.0\textwidth]{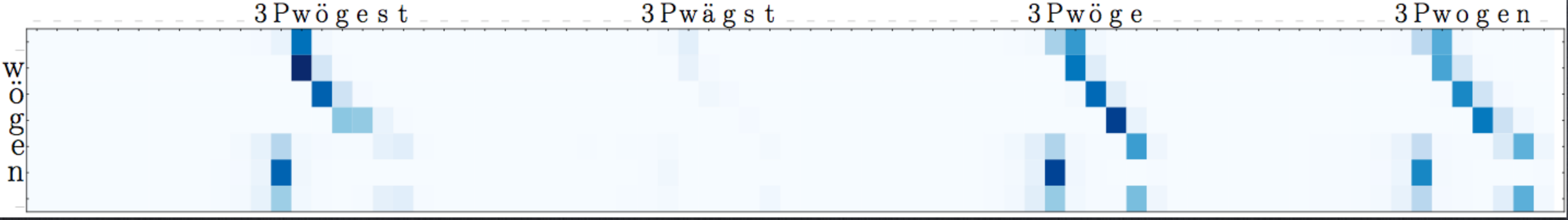}

 \caption{Attention heatmap for the multi-source  model. The
   example is for
   the German verb \word{wiegen} `to weigh'.
The model learns to focus most of its
   attention on forms that share the irregular subjunctive stem \word{w{\"o}g} in addition
   to the target subtags \word{3} and \word{P} that encode that the
   target form is 3rd person plural. We omit the tags from the diagram to which the model hardly attends.\label{fig:attention}}
\end{figure*}

\subsection{Attention Visualization}
\figref{attention} shows for one example, the generation of
the German form \word{w\"{o}gen}, \mtag{3rdPlSubPst}, the attention
weights of the multi-source model at each time step of the decoder,
i.e., for each character as it is being produced by the decoder. For
characters that simply need to be copied, the main attention lies
on the corresponding characters of the input sources. For example, the
character \word{g} is produced when attention is on the characters
\word{g} in \word{w\"{o}gest}, \word{w\"{o}ge} and \word{wogen}. This
aspect of the multi-source model is not different from
the single-source model, offering no advantage.

However, even for \word{g}, the source form that is least relevant for
generating \word{w\"{o}gen} receives almost no weight:
\word{w\"{a}gst} is an indicative singular form that does not provide
helpful information for generating a plural form in the subjunctive;
the model seems to have learned that this is the case. In contrast,
\word{wogen} does receive some weight; this makes sense as it is a
past indicative form and the past subjunctive is systematically
related to the past indicative for many German verbs. These
observations suggest that the network has learned to correctly predict
(at least in this case) which forms provide potentially useful
information.  For the last two time steps (i.e., characters to be
generated), attention is mainly focused on the tags. Again, this
indicates that the model has learned the regularity 
in generating this part of the word form: the suffix, consisting of
\word{en}, is predictable from the tag.

\section{Related Work}
\seclabel{RelWork}
Recently, variants of the RNN encoder-decoder have seen widespread
adoption in many areas of NLP due to their strong performance.
Encoder-decoders with and without attention have been applied to tasks
such as machine translation
\cite{cho2014properties,sutskever2014sequence,bahdanau2014neural},
parsing \cite{vinyals2015grammar} and automatic speech recognition
\cite{graves2005framewise,graves2013speech}.

The first work on multi-source models was presented for machine
translation. \newcite{zoph2016multi} made simultaneous use of source
sentences in multiple languages in order to find the best match
possible in the target language. Unlike our model, they apply
transformations to the hidden states of the encoders that are input to
the decoder.  \newcite{DBLP:journals/corr/FiratCB16}'s neural
architecture for MT translates from any of $N$ source languages to any
of $M$ target languages, using language specific encoders and
decoders, but sharing one single attention-mechanism.  In contrast to
our work, they obtain a single output for each input.

Much ink has been spilled on morphological reinflection over recent
years. \newcite{Dreyer_latent-variablemodeling} develop a
high-performing weighted finite-state transducer for the task, which was
later hybridized with an LSTM \cite{rastogi2016weighting}.
\newcite{durrett2013supervised} apply a semi-CRF to heuristically
extracted rules to generate inflected forms from lemmata using data
scraped from Wiktionary. Improved systems for the Wiktionary data
were subsequently developed by \newcite{mhulden2014}, who used a
semi-supervised approach, and \newcite{FaruquiTND15}, who used a
character-level LSTM. All of the above work has focused on the
single input case.  Two important exceptions, however, have
considered the multi-input case. Both \newcite{dreyer2009} and
\newcite{cotterell2015} define a string-valued graphical model over
the paradigm and apply the missing values. 

The SIGMORPHON 2016 Shared Task on Morphological Reinflection
\cite{cotterell-sigmorphon2016}, based on the {\sc UniMorph}
\cite{sylak-glassmankirov2015a} data, resulted in the development
of numerous  methods. RNN encoder-decoder
models
\cite{aharoni-goldberg-belinkov:2016:SIGMORPHON,kann16sigmorphon,ostling:2016:SIGMORPHON}
obtained the strongest performance and are the current state of the
art on the task. The best-performing model made use of an attention
mechanism \cite{kann16sigmorphon}, first popularized in machine translation \cite{bahdanau2014neural}.
We generalize this architecture to the multi-source case in this paper for the reinflection
task.

Besides generation, computational work on morphology has also focused
on analysis.  In this area, a common task---morphological
segmentation---is to break up a word into its sequence of constituent
morphs.  The unsupervised {\sc Morfessor} model
\cite{creutz2002unsupervised} has achieved widespread
adoption. Bayesian methods have also proven themselves successful in
unsupervised morphological segmentation \cite{johnson2006adaptor,goldwater2009bayesian}.
When labeled training data for segmentation is available, supervised methods
significantly outperform the unsupervised techniques \cite{ruokolainen2013supervised,cotterell-EtAl:2015:CoNLL,cotterell-vieira-schutze:2016:N16-1}.

As we pointed out in \secref{msr},
morphologically annotated corpora
provide an ideal source
of data for the multi-source MRI task: they
are annotated 
on the token level with inflectional
features and often contain several different inflected forms of
a lemma. 
\newcite{eskander-habash-rambow:2013:EMNLP}
develop an algorithm for automatic  learning of inflectional classes and
associated lemmas from morphologically annotated  corpora,
an approach that could be usefully combined with our
multi-source MRI framework.

\section{Conclusion}
Generation of unknown inflections in morphologically rich languages is
an important task that remains unsolved. We provide a new angle on the
problem by considering systems that are allowed to have multiple
inflected forms as input.  To this end, we define the task of multi-source
morphological reinflection as a generalization of single-source MRI
\cite{cotterell-sigmorphon2016} and present a model that solves the
task. We extend an attention-based RNN encoder-decoder architecture
from the single-source case to the multi-source case.  Our new model
consists of multiple encoders, each receiving one of the inputs. Our model improves over the state of the art for seven out
of eight languages, demonstrating the promise of multi-source MRI.
Additionally, we publically release our 
implementation.\footnote{\url{http://cistern.cis.lmu.de}}

\section{Future Work}
The new dataset for multi-source
morphological reinflection that 
we release
is a superset of the dataset of the
SIGMORPHON 2016 Shared Task on Morphological Reinflection to
facilitate research on morphological generation.  One focus of future
work should be the construction of more complex datasets, e.g.,
datasets that have better coverage of irregular words and datasets in
which there is no overlap in lemmata between training and test sets.
Further, for difficult inflections, it might be interesting to find an
effective way to include unsupervised data into the setup. For
example, we could define one of our $k$ inputs to be a form mined from
a corpus that is not guaranteed to have been correctly tagged
morphologically, but likely to be helpful.

We show in this paper that multi-source MRI outperforms single-source
MRI. This is an important contribution because---as we discussed in
\secref{motivate}---multi-source MRI is only promising for paradigms
with specific properties, which we referred to as {\sc SingleForm} and
{\sc MultiForm} configurations. Whether such configurations occur and
whether these configurations have a strong effect on MRI performance
was an open empirical question. Indeed, we found that for one of the
languages we investigated, for Hungarian, single-source MRI works at
least as well as multi-source MRI---presumably because its paradigms
almost exclusively contain {\sc SingleForm} configurations.  Thus,
single-source MRI is probably preferable for Hungarain since
single-source is simpler than multi-source.

There is another important question that we have not answered in this
paper: in an experimental setting in which the amount of training
information available is exactly the same for single-source and
multi-source, does multi-source still outperform single-source and by
how much? For example, the numbers we compare in \tabref{multi-source}
are matched with respect to the number of target forms, but not with
respect to the number of source forms: multi-source has more source
forms available for training than single-source. We leave
investigation of this important issue for future work.

\section*{Acknowledgments}
We  gratefully  acknowledge  the  financial  support
of Siemens and of DFG (SCHUE 2246/10-1) for this research.
The second author was supported by a DAAD Long-Term Research
Grant and an NDSEG fellowship.

\bibliographystyle{eacl2017}
\bibliography{multi-source-reinflection}

\begin{thebibliography}{}

\bibitem[\protect\citename{Aharoni \bgroup et al.\egroup
  }2016]{aharoni-goldberg-belinkov:2016:SIGMORPHON}
Roee Aharoni, Yoav Goldberg, and Yonatan Belinkov.
\newblock 2016.
\newblock Improving sequence to sequence learning for morphological inflection
  generation: The {BIU-MIT} systems for the {SIGMORPHON} 2016 shared task for
  morphological reinflection.
\newblock In {\em Proceedings of the 14th SIGMORPHON Workshop on Computational
  Research in Phonetics, Phonology, and Morphology}, pages 41--48, Berlin,
  Germany, August. Association for Computational Linguistics.

\bibitem[\protect\citename{Ahlberg \bgroup et al.\egroup
  }2015]{ahlberg2015paradigm}
Malin Ahlberg, Markus Forsberg, and Mans Hulden.
\newblock 2015.
\newblock Paradigm classification in supervised learning of morphology.
\newblock In {\em Proceedings of the 2015 Conference of the North American
  Chapter of the Association for Computational Linguistics: Human Language
  Technologies}, pages 1024--1029, Denver, Colorado, May--June. Association for
  Computational Linguistics.

\bibitem[\protect\citename{Bahdanau \bgroup et al.\egroup
  }2015]{bahdanau2014neural}
Dzmitry Bahdanau, Kyunghyun Cho, and Yoshua Bengio.
\newblock 2015.
\newblock Neural machine translation by jointly learning to align and
  translate.
\newblock In {\em Proceedings of the International Conference on Learning
  Representations}, San Diego, California, USA, May.

\bibitem[\protect\citename{Ballesteros \bgroup et al.\egroup
  }2015]{ballesteros2015improved}
Miguel Ballesteros, Chris Dyer, and Noah~A. Smith.
\newblock 2015.
\newblock Improved transition-based parsing by modeling characters instead of
  words with {LSTM}s.
\newblock In {\em Proceedings of the 2015 Conference on Empirical Methods in
  Natural Language Processing}, pages 349--359, Lisbon, Portugal, September.
  Association for Computational Linguistics.

\bibitem[\protect\citename{Bojar \bgroup et al.\egroup }2016]{bojarfindings2}
Ond\v{r}ej Bojar, Rajen Chatterjee, Christian Federmann, Yvette Graham, Barry
  Haddow, Matthias Huck, Antonio Jimeno~Yepes, Philipp Koehn, Varvara
  Logacheva, Christof Monz, Matteo Negri, Aurelie Neveol, Mariana Neves, Martin
  Popel, Matt Post, Raphael Rubino, Carolina Scarton, Lucia Specia, Marco
  Turchi, Karin Verspoor, and Marcos Zampieri.
\newblock 2016.
\newblock Findings of the 2016 conference on machine translation.
\newblock In {\em Proceedings of the First Conference on Machine Translation},
  pages 131--198, Berlin, Germany, August. Association for Computational
  Linguistics.

\bibitem[\protect\citename{Cho \bgroup et al.\egroup }2014]{cho2014properties}
Kyunghyun Cho, Bart van Merri\"{e}nboer, Dzmitry Bahdanau, and Yoshua Bengio.
\newblock 2014.
\newblock On the properties of neural machine translation: Encoder--decoder
  approaches.
\newblock In {\em Proceedings of SSST-8, Eighth Workshop on Syntax, Semantics
  and Structure in Statistical Translation}, pages 103--111, Doha, Qatar,
  October. Association for Computational Linguistics.

\bibitem[\protect\citename{Chrupa{\l}a}2008]{chrupala2008towards}
Grzegorz Chrupa{\l}a.
\newblock 2008.
\newblock {\em Towards a machine-learning architecture for lexical functional
  grammar parsing}.
\newblock {Ph.D.} thesis, Dublin City University.

\bibitem[\protect\citename{Cotterell \bgroup et al.\egroup
  }2015a]{cotterell-EtAl:2015:CoNLL}
Ryan Cotterell, Thomas M\"{u}ller, Alexander Fraser, and Hinrich Sch\"{u}tze.
\newblock 2015a.
\newblock Labeled morphological segmentation with semi-markov models.
\newblock In {\em Proceedings of the Nineteenth Conference on Computational
  Natural Language Learning}, pages 164--174, Beijing, China, July. Association
  for Computational Linguistics.

\bibitem[\protect\citename{Cotterell \bgroup et al.\egroup
  }2015b]{cotterell2015}
Ryan Cotterell, Nanyun Peng, and Jason Eisner.
\newblock 2015b.
\newblock Modeling word forms using latent underlying morphs and phonology.
\newblock {\em Transactions of the Association for Computational Linguistics},
  3:433--447.

\bibitem[\protect\citename{Cotterell \bgroup et al.\egroup
  }2016a]{cotterell-sigmorphon2016}
Ryan Cotterell, Christo Kirov, John Sylak-Glassman, David Yarowsky, Jason
  Eisner, and Mans Hulden.
\newblock 2016a.
\newblock The {SIGMORPHON} 2016 shared task---morphological reinflection.
\newblock In {\em Proceedings of the 14th SIGMORPHON Workshop on Computational
  Research in Phonetics, Phonology, and Morphology}, pages 10--22, Berlin,
  Germany, August. Association for Computational Linguistics.

\bibitem[\protect\citename{Cotterell \bgroup et al.\egroup
  }2016b]{cotterell-vieira-schutze:2016:N16-1}
Ryan Cotterell, Tim Vieira, and Hinrich Sch\"{u}tze.
\newblock 2016b.
\newblock A joint model of orthography and morphological segmentation.
\newblock In {\em Proceedings of the 2016 Conference of the North American
  Chapter of the Association for Computational Linguistics: Human Language
  Technologies}, pages 664--669, San Diego, California, June. Association for
  Computational Linguistics.

\bibitem[\protect\citename{Creutz and Lagus}2002]{creutz2002unsupervised}
Mathias Creutz and Krista Lagus.
\newblock 2002.
\newblock Unsupervised discovery of morphemes.
\newblock In {\em Proceedings of the ACL-02 Workshop on Morphological and
  Phonological Learning}, pages 21--30. Association for Computational
  Linguistics, July.

\bibitem[\protect\citename{Dreyer and Eisner}2009]{dreyer2009}
Markus Dreyer and Jason Eisner.
\newblock 2009.
\newblock Graphical models over multiple strings.
\newblock In {\em Proceedings of the 2009 Conference on Empirical Methods in
  Natural Language Processing}, pages 101--110, Singapore, August. Association
  for Computational Linguistics.

\bibitem[\protect\citename{Dreyer \bgroup et al.\egroup
  }2008]{Dreyer_latent-variablemodeling}
Markus Dreyer, Jason Smith, and Jason Eisner.
\newblock 2008.
\newblock Latent-variable modeling of string transductions with finite-state
  methods.
\newblock In {\em Proceedings of the 2008 Conference on Empirical Methods in
  Natural Language Processing}, pages 1080--1089, Honolulu, Hawaii, October.
  Association for Computational Linguistics.

\bibitem[\protect\citename{Dreyer}2011]{dreyer2011non}
Markus Dreyer.
\newblock 2011.
\newblock {\em A non-parametric model for the discovery of inflectional
  paradigms from plain text using graphical models over strings}.
\newblock {Ph.D.} thesis, Johns Hopkins University, Baltimore, MD.

\bibitem[\protect\citename{Durrett and DeNero}2013]{durrett2013supervised}
Greg Durrett and John DeNero.
\newblock 2013.
\newblock Supervised learning of complete morphological paradigms.
\newblock In {\em Proceedings of the 2013 Conference of the North American
  Chapter of the Association for Computational Linguistics: Human Language
  Technologies}, pages 1185--1195, Atlanta, Georgia, June. Association for
  Computational Linguistics.

\bibitem[\protect\citename{Eskander \bgroup et al.\egroup
  }2013]{eskander-habash-rambow:2013:EMNLP}
Ramy Eskander, Nizar Habash, and Owen Rambow.
\newblock 2013.
\newblock Automatic extraction of morphological lexicons from morphologically
  annotated corpora.
\newblock In {\em Proceedings of the 2013 Conference on Empirical Methods in
  Natural Language Processing}, pages 1032--1043, Seattle, Washington, USA,
  October. Association for Computational Linguistics.

\bibitem[\protect\citename{Faruqui \bgroup et al.\egroup }2016]{FaruquiTND15}
Manaal Faruqui, Yulia Tsvetkov, Graham Neubig, and Chris Dyer.
\newblock 2016.
\newblock Morphological inflection generation using character sequence to
  sequence learning.
\newblock In {\em Proceedings of the 2016 Conference of the North American
  Chapter of the Association for Computational Linguistics: Human Language
  Technologies}, pages 634--643, San Diego, California, June. Association for
  Computational Linguistics.

\bibitem[\protect\citename{Finkel and Stump}2007]{finkel2007principal}
Raphael Finkel and Gregory Stump.
\newblock 2007.
\newblock Principal parts and morphological typology.
\newblock {\em Morphology}, 17(1):39--75.

\bibitem[\protect\citename{Firat \bgroup et al.\egroup
  }2016]{DBLP:journals/corr/FiratCB16}
Orhan Firat, Kyunghyun Cho, and Yoshua Bengio.
\newblock 2016.
\newblock Multi-way, multilingual neural machine translation with a shared
  attention mechanism.
\newblock In {\em Proceedings of the 2016 Conference of the North American
  Chapter of the Association for Computational Linguistics: Human Language
  Technologies}, pages 866--875, San Diego, California, June. Association for
  Computational Linguistics.

\bibitem[\protect\citename{Goldwater \bgroup et al.\egroup
  }2009]{goldwater2009bayesian}
Sharon Goldwater, Thomas~L. Griffiths, and Mark Johnson.
\newblock 2009.
\newblock A bayesian framework for word segmentation: Exploring the effects of
  context.
\newblock {\em Cognition}, 112(1):21--54.

\bibitem[\protect\citename{Graves and Schmidhuber}2005]{graves2005framewise}
Alex Graves and J{\"u}rgen Schmidhuber.
\newblock 2005.
\newblock Framewise phoneme classification with bidirectional {LSTM} and other
  neural network architectures.
\newblock {\em Neural Networks}, 18(5):602--610.

\bibitem[\protect\citename{Graves \bgroup et al.\egroup
  }2013]{graves2013speech}
Alex Graves, Abdel{-}rahman Mohamed, and Geoffrey~E. Hinton.
\newblock 2013.
\newblock Speech recognition with deep recurrent neural networks.
\newblock In {\em {IEEE} International Conference on Acoustics, Speech and
  Signal Processing}, pages 6645--6649, Vancouver, BC, Canada, May.

\bibitem[\protect\citename{Hulden \bgroup et al.\egroup }2014]{mhulden2014}
Mans Hulden, Markus Forsberg, and Malin Ahlberg.
\newblock 2014.
\newblock Semi-supervised learning of morphological paradigms and lexicons.
\newblock In {\em Proceedings of the 14th Conference of the European Chapter of
  the Association for Computational Linguistics}, pages 569--578, Gothenburg,
  Sweden, April. Association for Computational Linguistics.

\bibitem[\protect\citename{Johnson \bgroup et al.\egroup
  }2006]{johnson2006adaptor}
Mark Johnson, Thomas~L. Griffiths, and Sharon Goldwater.
\newblock 2006.
\newblock Adaptor grammars: {A} framework for specifying compositional
  nonparametric bayesian models.
\newblock In {\em Advances in Neural Information Processing Systems 19}, pages
  641--648, Vancouver, BC, Canada, December.

\bibitem[\protect\citename{Kann and Sch\"{u}tze}2016a]{kann16sigmorphon}
Katharina Kann and Hinrich Sch\"{u}tze.
\newblock 2016a.
\newblock {MED}: The {LMU} system for the {SIGMORPHON} 2016 shared task on
  morphological reinflection.
\newblock In {\em Proceedings of the 14th SIGMORPHON Workshop on Computational
  Research in Phonetics, Phonology, and Morphology}, pages 62--70, Berlin,
  Germany, August. Association for Computational Linguistics.

\bibitem[\protect\citename{Kann and Sch\"{u}tze}2016b]{kann16singlemodel}
Katharina Kann and Hinrich Sch\"{u}tze.
\newblock 2016b.
\newblock Single-model encoder-decoder with explicit morphological
  representation for reinflection.
\newblock In {\em Proceedings of the 54th Annual Meeting of the Association for
  Computational Linguistics (Volume 2: Short Papers)}, pages 555--560, Berlin,
  Germany, August. Association for Computational Linguistics.

\bibitem[\protect\citename{Le \bgroup et al.\egroup }2015]{le2015simple}
Quoc~V. Le, Navdeep Jaitly, and Geoffrey~E. Hinton.
\newblock 2015.
\newblock A simple way to initialize recurrent networks of rectified linear
  units.
\newblock {\em CoRR}, abs/1504.00941.

\bibitem[\protect\citename{McDonald \bgroup et al.\egroup
  }2013]{mcdonald2013universal}
Ryan McDonald, Joakim Nivre, Yvonne Quirmbach-Brundage, Yoav Goldberg, Dipanjan
  Das, Kuzman Ganchev, Keith Hall, Slav Petrov, Hao Zhang, Oscar
  T\"{a}ckstr\"{o}m, Claudia Bedini, N\'{u}ria Bertomeu~Castell\'{o}, and
  Jungmee Lee.
\newblock 2013.
\newblock Universal dependency annotation for multilingual parsing.
\newblock In {\em Proceedings of the 51st Annual Meeting of the Association for
  Computational Linguistics (Volume 2: Short Papers)}, pages 92--97, Sofia,
  Bulgaria, August. Association for Computational Linguistics.

\bibitem[\protect\citename{M{\"{u}}ller \bgroup et al.\egroup
  }2015]{thomasjoint}
Thomas M{\"{u}}ller, Ryan Cotterell, Alexander~M. Fraser, and Hinrich
  Sch{\"{u}}tze.
\newblock 2015.
\newblock Joint lemmatization and morphological tagging with lemming.
\newblock In {\em Proceedings of the 2015 Conference on Empirical Methods in
  Natural Language Processing}, pages 2268--2274, Lisbon, Portugal, September.
  Association for Computational Linguistics.

\bibitem[\protect\citename{Nicolai \bgroup et al.\egroup
  }2015]{nicolai2015inflection}
Garrett Nicolai, Colin Cherry, and Grzegorz Kondrak.
\newblock 2015.
\newblock Inflection generation as discriminative string transduction.
\newblock In {\em Proceedings of the 2015 Conference of the North American
  Chapter of the Association for Computational Linguistics: Human Language
  Technologies}, pages 922--931, Denver, Colorado, May--June. Association for
  Computational Linguistics.

\bibitem[\protect\citename{\"{O}stling}2016]{ostling:2016:SIGMORPHON}
Robert \"{O}stling.
\newblock 2016.
\newblock Morphological reinflection with convolutional neural networks.
\newblock In {\em Proceedings of the 14th SIGMORPHON Workshop on Computational
  Research in Phonetics, Phonology, and Morphology}, pages 23--26, Berlin,
  Germany, August. Association for Computational Linguistics.

\bibitem[\protect\citename{Rastogi \bgroup et al.\egroup
  }2016]{rastogi2016weighting}
Pushpendre Rastogi, Ryan Cotterell, and Jason Eisner.
\newblock 2016.
\newblock Weighting finite-state transductions with neural context.
\newblock In {\em Proceedings of the 2016 Conference of the North American
  Chapter of the Association for Computational Linguistics: Human Language
  Technologies}, pages 623--633, San Diego, California, June. Association for
  Computational Linguistics.

\bibitem[\protect\citename{Ruokolainen \bgroup et al.\egroup
  }2013]{ruokolainen2013supervised}
Teemu Ruokolainen, Oskar Kohonen, Sami Virpioja, and Mikko Kurimo.
\newblock 2013.
\newblock Supervised morphological segmentation in a low-resource learning
  setting using conditional random fields.
\newblock In {\em Proceedings of the Seventeenth Conference on Computational
  Natural Language Learning}, pages 29--37, Sofia, Bulgaria, August.
  Association for Computational Linguistics.

\bibitem[\protect\citename{Stump and Finkel}2013]{stump2013morphological}
Gregory Stump and Raphael~A. Finkel.
\newblock 2013.
\newblock {\em Morphological typology: From word to paradigm}, volume 138.
\newblock Cambridge University Press.

\bibitem[\protect\citename{Sutskever \bgroup et al.\egroup
  }2014]{sutskever2014sequence}
Ilya Sutskever, Oriol Vinyals, and Quoc~V. Le.
\newblock 2014.
\newblock Sequence to sequence learning with neural networks.
\newblock In {\em Advances in Neural Information Processing Systems 27}, pages
  3104--3112, Montreal, Quebec, Canada, December.

\bibitem[\protect\citename{Sylak-Glassman \bgroup et al.\egroup
  }2015]{sylak-glassmankirov2015a}
John Sylak-Glassman, Christo Kirov, David Yarowsky, and Roger Que.
\newblock 2015.
\newblock A language-independent feature schema for inflectional morphology.
\newblock In {\em Proceedings of the 53rd Annual Meeting of the Association for
  Computational Linguistics and the 7th International Joint Conference on
  Natural Language Processing (Volume 2: Short Papers)}, pages 674--680,
  Beijing, China, July. Association for Computational Linguistics.

\bibitem[\protect\citename{Vinyals \bgroup et al.\egroup
  }2014]{vinyals2015grammar}
Oriol Vinyals, Lukasz Kaiser, Terry Koo, Slav Petrov, Ilya Sutskever, and
  Geoffrey~E. Hinton.
\newblock 2014.
\newblock Grammar as a foreign language.
\newblock {\em CoRR}, abs/1412.7449.

\bibitem[\protect\citename{Zeiler}2012]{zeiler2012adadelta}
Matthew~D Zeiler.
\newblock 2012.
\newblock Adadelta: an adaptive learning rate method.
\newblock {\em arXiv preprint arXiv:1212.5701}.

\bibitem[\protect\citename{Zoph and Knight}2016]{zoph2016multi}
Barret Zoph and Kevin Knight.
\newblock 2016.
\newblock Multi-source neural translation.
\newblock In {\em Proceedings of the 2016 Conference of the North American
  Chapter of the Association for Computational Linguistics: Human Language
  Technologies}, pages 30--34, San Diego, California, June. Association for
  Computational Linguistics.

\end{thebibliography}

\end{document}